\documentclass[11pt,a4paper]{article}
\usepackage[hyperref]{eacl2021}
\usepackage{times}
\usepackage{latexsym}
\usepackage[T1]{fontenc}
\usepackage[utf8]{inputenc}
\usepackage{tabularx}
\usepackage{array}
\usepackage{url}
\usepackage{graphicx}
\usepackage{multirow}
\usepackage[inline]{enumitem}
\usepackage{booktabs,subcaption,amsfonts,dcolumn}
\usepackage{cleveref}

\newcommand{\PreserveBackslash}[1]{\let\temp=\\#1\let\\=\temp}
\newcolumntype{R}[1]{>{\PreserveBackslash\raggedleft}p{#1}}

\usepackage{microtype}

\aclfinalcopy 
\def\aclpaperid{79} 

\title{The Gutenberg Dialogue Dataset}

\author{Richard Csaky \\
  Budapest University of Technology \\
  University of Oxford \\
  \texttt{richard.csaky@psych.ox.ac.uk} \\\And
  G\'{a}bor Recski \\
  TU Wien \\
  \texttt{gabor.recski@tuwien.ac.at} \\}

\date{}

\begin{document}
\maketitle
\begin{abstract}
Large datasets are essential for neural modeling of many NLP tasks.
Current publicly available open-domain dialogue datasets offer a
trade-off between quality (e.g.,~DailyDialog \cite{Li:2017b}) and size
(e.g.,~Opensubtitles
\cite{Tiedemann:2012}). We narrow this gap by building a high-quality
dataset of 14.8M utterances in English, and smaller datasets in German,
Dutch, Spanish, Portuguese, Italian, and Hungarian.  We extract and
process dialogues from public-domain books made available by Project Gutenberg\footnote{\url{https://www.gutenberg.org/}}.
We describe our dialogue extraction pipeline,
analyze the effects of the various heuristics used, and present an
error analysis of extracted dialogues. Finally, we conduct experiments
showing that better response quality can be achieved in zero-shot and
finetuning settings by training on our data than on the larger but much
noisier Opensubtitles dataset.  Our open-source
pipeline\footnote{\url{https://github.com/ricsinaruto/gutenberg-dialog}} can be extended to further
languages with little additional effort.  Researchers can also build
their versions of existing datasets by adjusting various trade-off
parameters.
\end{abstract}
    
\section{Introduction}
\label{sec:introduction}
Current open-domain dialogue datasets offer trade-offs between quality and
size. High-quality datasets are usually too small to represent the multitude of
topics required for a conversational agent. Large datasets often lack good
turn-segmentation and are generally noisy, models trained on such datasets
generate low-quality or generic output. In \Cref{sec:background} we analyze
publicly available dialogue corpora and the trade-offs they offer.  To address
the need for large, high-quality datasets we build a corpus of 14.8M utterances
in English using publicly available books from Project
Gutenberg. We also build datasets
for German, Dutch, Spanish, Portuguese, Italian, and Hungarian, with utterance
counts in the 20k--200k range.  We call this dataset ensemble the Gutenberg
Dialogue Dataset. We wish to make it explicit that we are not aiming to create a gold dataset. Our goal is to create a dataset which offers a better size-quality trade-off than other dialogue corpora. The Gutenberg dataset is both larger than DailyDialog \cite{Li:2017b} and has better quality than Opensubtitles \cite{Tiedemann:2012}, and we think it benefits researchers by filling a need in the landscape of dialogue corpora. The Gutenberg Dialogue Dataset and the code used to build it can be accessed through this repository: \url{https://github.com/ricsinaruto/gutenberg-dialog}. The repository also contains all trained models presented in this paper and all data and training scripts used to produce the results. We also built a web demo interface for interacting with the trained models\footnote{\url{https://ricsinaruto.github.io/chatbot.html}}.

In \Cref{sec:dataset} we offer a detailed quantitative
analysis of our heuristics to better understand their effects on data quality.
\Cref{ssec:errors} presents our error analysis of the English
dataset both at the utterance and dialogue level.  Using our MIT licensed
pipeline, researchers can easily build various dataset versions by adjusting
a small number of parameters that control multiple dimensions of the
size-quality trade-off.  In \Cref{sec:trainings} we evaluate our
dataset in a generative multi-turn and single-turn setting using the GPT-2
\cite{Radford:2019} and Transformer \cite{Vaswani:2017} architectures,
respectively. For each of the 7 languages, we compare models trained on
Gutenberg and Opensubtitles. For English, we also compare
zero-shot and finetuning performance of Gutenberg and Opensubtitles on two
smaller datasets.  Potential improvements and future work is discussed in
\Cref{sec:conclusion}.  Extension to additional languages is ongoing, we
welcome all contributions from the community: our modular code requires only a
limited amount of language-specific effort for each new language.

\section{Background}
\label{sec:background}
    
Open-domain dialogue datasets vary in size, quality, and source, as
demonstrated in \Cref{table:datasets}. Generally, smaller datasets are constructed using controlled crowdsourcing environments, making their quality higher
(e.g.,~PersonaChat \cite{Zhang:2018}). Crowdsourcing platforms like Amazon
Mechanical Turk\footnote{\url{https://www.mturk.com/}} are used to hire and instruct workers to carry out free-form conversations.
    Larger datasets can be built by automatic processing of dialogue-like text
    sources, such as Opensubtitles and
    Reddit\footnote{\url{https://www.reddit.com/}} \cite{Henderson:2019}).
    Opensubtitles contains movie subtitles in multiple languages and Reddit is a discussion forum with millions of daily comments on various topics. Automatic extraction offers less quality control, and the data source heavily influences the genre of conversations. In Reddit data, everyday chit-chat is less common, comments in the same thread all discuss the same post. Two-party dialogues are rare as threads are almost always multi-speaker. Twitter\footnote{\url{https://twitter.com/}} conversations have similar problems and they are also constrained by a character limit. Extracting conversations from Twitter and Reddit is straightforward as speaker segmentation is included and the thread chain can be used as dialogue history.

    Books, especially fiction, have so far seen little use as a
    source of dialogue data.
    In DailyDialog \cite{Li:2017b}, 90 000 high-quality utterances are
    extracted from online resources for English language learners,
    extraction steps are not detailed. The quality of these dialogues and the lack of a large book-based dataset motivates our work. Dialogues extracted from books, like movie subtitles,
    lack context, but their usefulness is evidenced by the Cornell Corpus \cite{Danescu:2011} and
    DailyDialog. As argued by \citet{Danescu:2011} and \citet{Fainberg:2018},
    artificial dialogues in movies and books generally resemble natural
    conversations. Such dialogues are also called written dialogues as opposed to spoken corpora like the Switcboard corpus \cite{Godfrey:1992}. Though our corpus contains written dialogues we also perform evaluation on Persona-Chat, which can be considered as a spoken dialogue corpus, and show Gutenberg's effectiveness in this setting as well.
    
    Unfortunately, the Cornell Corpus is
    relatively small, while the Opensubtitles corpus suffers from the fact that
    the original dataset lacks both dialogue and turn segmentation:
    subtitle lines
    are treated as turns and dialogue history consists of the previous
    \textit{n} lines, with little to no additional post-processing used to extract dialogues instead of using the raw data (\citet{Henderson:2019} removes the shortest and longest utterances to improve quality). These issues lead to trained models outputting generic responses, e.g.,~to the
    input ``yes i believe there are green teas black teas and scented teas. any
    others?'' a model trained on Opensubtitles outputs ``sure.''.
    In
    addition, the types and ratio of errors in these datasets have not been
    explicitly analyzed. For the Gutenberg dataset, we build a multi-step extraction
    pipeline and analyze both the performance of each heuristic and the ratio of
    each error type in a sample of the final corpus. Unfortunately, most of the tools developed here are specific to the book domain, and use textual patterns which are not available in Opensubtitles. In order to increase the quality of Opensubtitles, subtitle-specific methods need to be developed, like taking into account the elapsed time between two subtitle lines.
    
    The size of our corpus facilitates effective training of large
    Transformer-based models \cite{Radford:2019,Yang:2019}. Recently,
    pre-training and finetuning large language models on specific tasks
    (including dialogue modeling) has gained popularity
    \cite{Wolf:2019,Devlin:2018a}. Transformer-based models and specifically
    GPT-2 have gained state-of-the-art status in the dialogue domain
    \cite{Adiwardana:2020,Roller:2020,Zhang:2019d,Wolf:2019}. 
    Through these models the community has gradually shifted from single-turn
    to multi-turn scenarios. Since we wish to demonstrate our dataset's quality
    on the dialogue-level, we conduct experiments primarily with GPT-2. We report some
    single-turn trainings using Transformer for comparison. We show Gutenberg's
    effectiveness for multi-turn pre-training in \Cref{sec:trainings},
    comparing it to Opensubtitles pre-training, which is popular in the
    literature \cite{Csaky:2017,Krause:2017a,Xing:2018a}.

    \begin{table*}[h!]
        \begin{center}
            \renewcommand{\arraystretch}{1.0}
            \begin{tabular}{p{7cm}rll}
                \bf Dataset & \bfseries{Size} & \bf Source & \bf Quality   \\ 
                \midrule
                
                DailyDialog \cite{Li:2017b} & 90k & ESL websites & auto-extracted  \\ 
                
                Wizard-of-Wikipedia \cite{Dinan:2018} & 100k & crowdsourcing & human-written \\
                
                Document-grounded \cite{Zhou:2018b} & 100k & crowdsourcing & human-written \\  
                
                Persona-Chat \cite{Zhang:2018} & 150k & crowdsourcing & human-written \\ 
                
                Self-dialogue \cite{Fainberg:2018} & 150k & crowdsourcing & human-written \\ 
                
                Cornell Movie Corpus \cite{Danescu:2011} & 300k & movie scripts & auto-extracted \\ 
                Self-feeding chatbot \cite{Hancock:2019} & 500k & human-bot dialogues & partly human-written\\ 
                Twitter corpus\footnotemark{} & 5M & Twitter posts/replies & auto-extracted  \\ 
                Opensubtitles \cite{Henderson:2019} & 320M & movie subtitles & auto-extracted  \\ 
                
                Reddit \cite{Henderson:2019} & 730M & Reddit threads & auto-extracted  \\ 
            \end{tabular}
        \end{center}
        \caption{\label{table:datasets} Comparison of open-domain dialogue datasets in English. \textit{Size} is the rough number of utterances, \textit{Source} describes where the data comes from, and \textit{Quality} distinguishes between dataset collection techniques.}
    \end{table*}

    \section{Extraction Pipeline}
    \label{sec:dataset}
    Most of Project Gutenberg's 60 000 online books are in English (47 300
    books; 3 billion words). French, Finnish, and German, the next most common languages, contain 3000, 2000, 1750 books, and 194M, 74M, 82M
    words, respectively. Dutch, Spanish, Italian, Portuguese, and Chinese are
     \footnotetext[7]{\url{https://github.com/Marsan-Ma-zz/chat_corpus}}
    all above 10M words, followed by a long tail of various languages.  We used
    the Gutenberg python
    package\footnote{\url{https://github.com/ageitgey/Gutenberg}} to download
    books and query their license, language, and author metadata. Further Gutenberg statistics can be found in \Cref{ssec:stat_appendix}.
    This section describes heuristics and methods used to extract dialogues
    from books and remove noise. The main challenges are identifying changes
    between speakers within a dialogue and separating sets of utterances that
    do not belong to the same dialogue. To separate dialogues,
    changes in location, time, speaker, etc. would have to be identified directly, but we
    develop simple heuristics (e.g.,~distance between utterances) that
    can extract relatively high-quality conversations at scale. Tunable
    parameters of our system offer trade-offs between data quality and
    size. Using our open-source system researchers can build custom datasets that best suit their applications.
    
    Our dialogue extraction pipeline includes three main steps:
    \begin{enumerate*}
        \item Conversational and narrative text is separated.
        \item Dialogic text is split into separate dialogues. 
        \item Dialogues are segmented into separate turns (utterances).
    \end{enumerate*}
    In most books, conversational text is highlighted; e.g.,~placed between
    single/double quotation marks in English or started by an em-dash in
    Hungarian. Naturally, these delimiters have other uses as well, but such
    cases are rare (about 5\% of utterances, see \Cref{ssec:errors}).
    We can only extract dialogues from books which clearly delimit both the
    start and end of conversations. In some languages/books, the start of an
    utterance is given, but the end is not, and narrative text can get mixed in
    (e.g.,~\textit{Si vous arrivez avant nous, cria Luigi au messager, annoncez
    à la nourrice que nous vous suivons.} `If you arrive before us, shouted
    Luigi to the messenger, tell the nurse that we are following you.').
    This is why we could not build a French dataset, and have relatively
    smaller datasets in Dutch, Italian, Portuguese, and Hungarian.
    \Cref{figure:example} shows a sample dialogue highlighting our heuristics.
    In the following paragraphs, we offer a parameter-based description of our
    pipeline.
    
        \begin{figure}[h!]
        \small
        \begin{center}
            \begin{tabular}{p{7.5cm}}
                "Read what I have written," she gasped. "It may be utterly
                unintelligible."\\
                
                For answer, Morton folded the sheet and placed it in an envelope.\\
                
                "Address this, if you please," he said.\\
                
                She obeyed his request, limply forcing herself to make the effort;
                and, as the pen once more fell from her fingers, she glanced up at him with a haggard piteousness in her eyes.\\
                
                "Will you not read what I have written?" she asked again.\\
                
                "I see no reason why I should," he answered.\\

            \end{tabular}
        \end{center}
        \caption{\label{figure:example} A dialogue example. Utterances are in separate paragraphs, sometimes broken up by narrative text.}
    \end{figure}
    
     \begin{table*}[h!]
        \small
        \renewcommand{\arraystretch}{1.1}
        \begin{center}
            \begin{tabular}{llll}
                \bf Method & \bf Parameter & \bf Filtered & \bf What   \\ \midrule
                
                Pre-filter & 2 (KL-div) & 2090 books (4.42\%) & Old books and noise   \\ 
                
                Delimiter filter & 150 delimiters / 10 000 words & 20 500 books (43.3\%) & Books with no dialogues \\
                
                Long utterances & 100 words & 610 000 utterances (3.95\%)& Non-conversational utterances  \\ 
                
                Post-filter & 20\% rare words & 20 478 dialogues (0.8\%) & Dialogues containing many rare words \\ 

            \end{tabular}
        \end{center}
        \caption{\label{table:filterings} The various filtering steps for the English dataset.}
    \end{table*}
    
    \paragraph{Pre-filtering} After downloading books and separating them by language,
    all copyrighted works are removed. We also filter books containing
    unusual, mostly older, 
    language: if the KL divergence between a book's word distribution and the
    total (all books) distribution is above a threshold (2),
    it is removed. The method is less accurate for short books with less than
    20 000 words, these are not filtered. In the English dataset, 2090
    books were removed (4.42\%). By analyzing 100 filtered and 100 non-filtered
    books randomly, we found 8 false positives (books that should not have been
    removed), and 9 false negatives. 
    
    \paragraph{Delimiter filter} Before dialogue extraction, books with less
    than 150 delimiters per 10 000 words are removed. We assume that under a
    certain threshold the probability of delimiters used for non-conversational
    purposes is increased. We empirically set this ratio by increasing it until
    the assumption starts failing. Since many books do not contain dialogues,
    almost half were removed (20 500) in the English pipeline. Sampling 100
    filtered and 100 non-filtered books, we found 8 false positives (books that
    should not have been removed), and 22 false negatives.
    In a sample of the final dataset, less than 5\% of utterances
    were non-conversational (cf. \Cref{ssec:errors}).

    \paragraph{Dialogue gap} If two dialogue segments highlighted by delimiters
    are far apart, i.e.~there are >150 characters between them, they will not be
    considered part of the same dialogue. This heuristic, the dialogue gap,
    will always offer a false positive/negative trade-off since the amount of
    text between dialogues varies considerably. We tuned this trade-off by reasoning
    that shorter dialogues are less problematic than incoherent dialogues: our
    setting yields 3.5 times fewer false negatives, as shown in \Cref{ssec:errors}.
    Our turn segmentation heuristic will also always treat separate paragraphs
    as separate utterances. In a sample of the final dataset, this assumption
    fails for
    roughly 4\% of utterance pairs (cf. \Cref{ssec:errors}).

    \paragraph{Long utterances and rare words} During dialogue extraction utterances with more
    than 100 words are removed to ensure that remaining utterances are truly
    conversational and to facilitate neural model training \cite{Dai:2019}. As
    all other parameters in the pipeline, this is adjustable to the needs of
    the user or task.
    Finally, we remove dialogues with more than 20\% rare words
    (not in the top 100 000), removing noise and facilitating neural model
    training.
    Dialogues are split randomly into train (90\%), validation
    (5\%), and test (5\%) datasets, dialogues from the same book are placed in
    the same split.
    
         \begin{table}[ht!]
        \small
        \renewcommand{\arraystretch}{1.1}
        \begin{center}
            \begin{tabular}{lrrrr}
                &  \#U &  \(|U|\) &  \#D  &  \(|D|\)   \\ \midrule
                
                \bf English & 14 773 741 & 22.17 & 2 526 877 & 5.85  \\ 
                
                \bf German & 226 015 & 24.44 & 43 440 & 5.20 \\ 
                
                \bf Dutch & 129 471 & 24.26 & 23 541 & 5.50 \\  
                
                \bf Spanish & 58 174 & 18.62 & 6 912 & 8.42 \\ 
                
                \bf Italian & 41 388 & 19.47 & 6 664 & 6.21 \\ 
                
                \bf Hungarian & 18 816 & 14.68 & 2 826 & 6.66 \\
                
                \bf Portuguese & 16 228 & 21.40 & 2 233 & 7.27 \\
                
            \end{tabular}
        \end{center}
        \caption{\label{table:stats} Statistics of the final dialogue datasets. Columns are: language, number of utterances, average utterance length, number of dialogues, and average dialogue length.}
    \end{table}

    Languages differ only in the dialogue extraction step. The modular pipeline
    can be easily extended to new languages by specifying conversational
    delimiters and a minimal implementation of dialogue and turn segmentation,
    generally adaptable from English. In practice, adapting the English pipeline to other languages ranged between 0-50 lines of python code. Optionally further analysis might be needed to check the output of the pipeline and refine the extracting process if needed. Delimiters and parameters for other
    languages were not analyzed as profoundly as for English, leaving room for
    improvements in future work. We aim to show that good dialogue datasets can
    be constructed with minimal effort, as a first step towards a high-quality
    multi-language dataset ensemble. In total, the four filtering steps removed about 12.5\% of utterances from
    the English dataset, detailed in \Cref{table:filterings}. Statistics of
    the final datasets in all 7 languages can be seen in \Cref{table:stats}.
    The standard deviation of dialogue length in English is 6.09, and there are
    87 500 dialogues with at least 20 utterances. The average dialogue length
    can be linearly adjusted with the dialogue gap parameter.

    \section{Error Analysis}
    \label{ssec:errors}
    
     \paragraph{Utterance-level} To assess the single-turn quality of the
    English dataset we manually analyzed 100 random utterance pairs with book context. 89 pairs did not contain any errors. Remaining utterance pairs contained 1 error type each, out of 2 major and 2 minor types, minor errors occurring in only 1 case
    each. The extracted text is not conversational in 5 utterance pairs, a
    consequence of the delimiter threshold and other sources of noise (\Cref{figure:example_a}).
    Utterances of a single speaker were falsely treated as multiple turns in 4
    cases, most often because of our assumption that paragraph breaks signal
    dialogue turns (\Cref{figure:example_d}).
    
    		\begin{figure}[h!]
		\small
		\begin{center}
			\begin{tabular}{p{7.5cm}}
				And he was singing, too, as he went on with his task; sometimes--\\
				"Play on, minstrèl, play on, minstrèl,
				My lady is mine only girl;"\\
				
			\end{tabular}
		\end{center}
		\caption{\label{figure:example_a} Non-dialogue text detected as an utterance.}
	\end{figure}
	
		\begin{figure}[h!]
		\small
		\begin{center}
			\begin{tabular}{p{7.5cm}}
				In his progress he passed the door of the dormitory of his victim—he
				paused a moment, and listened attentively. Then in a voice of deep
				anguish he said,—\\
				
				“She can sleep—she can sleep—no ghostly vision scares slumber from her
				eyes—while—”\\
				
				He shuddered, and passed a step or two on, then pausing again, he said,—\\
				
				“Oh, if she, the young and innocent...”\\

			\end{tabular}
		\end{center}
		\caption{\label{figure:example_d} Two consecutive turns uttered by the same speaker.}
	\end{figure}

\begin{figure}[!ht]
        \centering
        \includegraphics[width=0.48\textwidth]{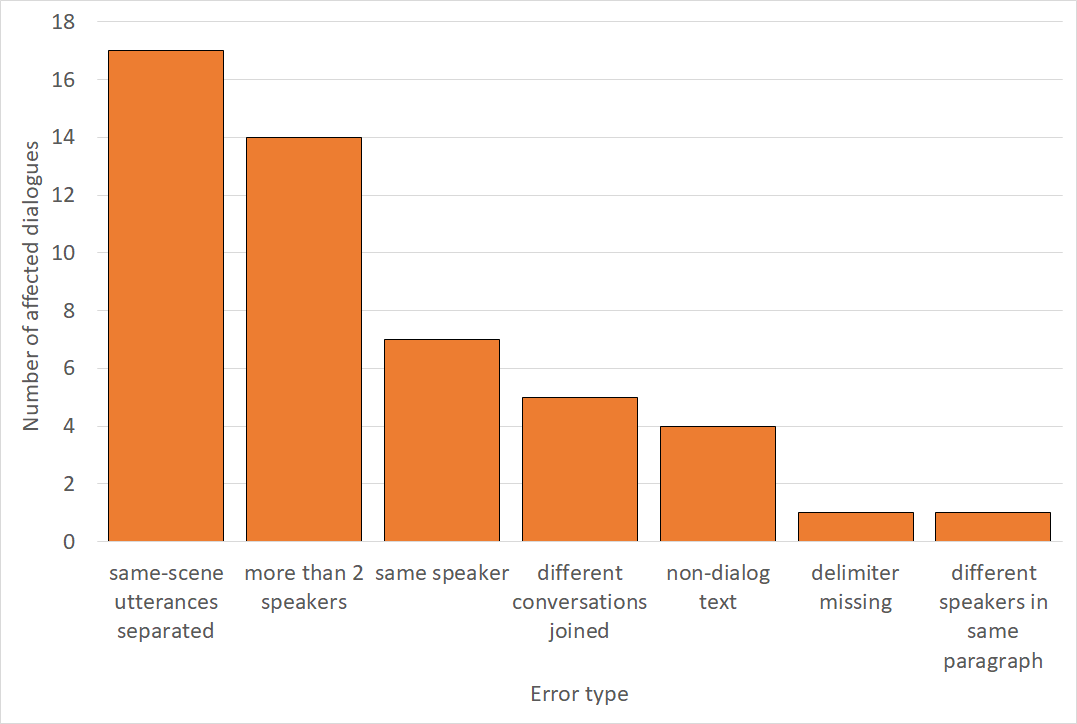}
        \caption{Number of dialogues affected by the various errors. In total 50 dialogues were analyzed. Some dialogues contained multiple types of errors and only 16 dialogues contained 0 errors.}
        \label{fig:delimiters}
    \end{figure}

    \paragraph{Dialogue-level} Errors in whole dialogues exhibit a much greater
    variety. Based on a manual analysis of 50 dialogues in the English
    dataset we identified 7 error categories (\Cref{fig:delimiters}). The following numbers are always out of the 50 analyzed dialogues. 16 dialogues contained 0 errors, 21 contained 1 error type, 11 contained 2
    types, remaining dialogues containing 3. We detail the number of dialogues
    affected by each error type below. We note that this does not constitute a
    proper statistical analysis.

    	\begin{figure}[h!]
		\small
		\begin{center}
			\begin{tabular}{p{7.5cm}}
				Richard curbed an impatient rejoinder, and said quietly, "William
				Durgin had an accomplice."\\
				
				Mr. Taggett flushed, as if Richard had read his secret thought. Durgin's flight, if he really had fled, had suggested a fresh
				possibility to Mr. Taggett. What if Durgin were merely the pliant
				instrument of the cleverer man who was now using him as a shield?
				This reflection was precisely in Mr. Taggett's line. In absconding
				Durgin had not only secured his own personal safety, but had
				exonerated his accomplice. It was a desperate step to take, but it
				was a skillful one.\\
				
				"He had an accomplice?" repeated Mr. Taggett, after a moment. "Who
				was it?\\

			\end{tabular}
		\end{center}
		\caption{\label{figure:example_c} A single conversation cut up because of the long paragraph between the two utterances.}
	\end{figure}

        \begin{figure}[h!]
		\small
		\begin{center}
			\begin{tabular}{p{7.5cm}}
				“Carry pins, is it?” said Tom. “Ye can carry yer head level, me boy. So at it ye go, an' ye'll bate Rory fer me, so ye will.”\\
				
				“Well then,” cried Barney, “I will, if you give me first choice, and
				I'll take Tom here.”\\
				
				“Hooray!” yelled Tom, “I'm wid ye.” So it was agreed, and in a few
				minutes the sides were chosen, little Ben Fallows falling to Rory as
				last choice.\\
				
				“We'll give ye Ben,” said Tom, whose nerve was coming back to him. “We don't want to hog on ye too much.”\\
				
				“Never you mind, Ben,” said Rory, as the little Englishman strutted to his place among Rory's men. “You'll earn your supper to-day with the best of them.”\\

			\end{tabular}
		\end{center}
		\caption{\label{figure:example_b} First three and last two utterances are not part of the same conversation, but they were merged because of the dialogue gap threshold.}
	\end{figure}
	
	 Utterances from the same conversation frequently end up in different
    dialogues (17 cases, example in \Cref{figure:example_c}) because of the dialogue gap threshold. The inverse, a dialogue containing utterances
    from multiple conversations, occurred in 5 cases (\Cref{figure:example_b}). While it is challenging to set this parameter, we
    consider this to be a reasonable trade-off: shorter dialogues mean
    less data,
    but incoherent dialogues with utterances from multiple conversations are
    bad data.
    In \Cref{sec:conclusion} we discuss possible further approaches to segmenting
    conversational text.

    Books often contain dialogues between more than two speakers, our second
    most frequent source of error (14 dialogues). However, such conversations are still
    coherent and provide useful data for model training. In contrast, the same
    speaker uttering at least two consecutive turns breaks coherence in 7
    dialogues. Tackling these issues would have to involve speaker
    identification (cf.
    \Cref{sec:conclusion}). As in the utterance-level analysis, there were some
    dialogues (4) in which non-conversational text got mixed in. The remaining errors, \textit{delimiter missing}
    and \textit{different speakers in same paragraph} occurred in only 1
    dialogue out of 50.

    \section{Experiments}
    \label{sec:trainings}
    \subsection{Evaluation Metrics}
    
    Most automatic evaluation methods 
    for dialogue models
    correlate poorly with human judgment \cite{Liu:2016}, and recently proposed
    metrics that correlate better \cite{Li:2017a,Lowe:2017,Tao:2018} are harder
    to measure than perplexity or BLEU \cite{Papineni:2002}. Human evaluation
    also has its shortcomings, like high variance, cost, and replication
    difficulty \cite{Zhang:2018,Tao:2018}. There does not seem to be any
    consensus on the best approach,    as some researchers use only automatic
    metrics \cite{Xing:2018a,Xu:2018}, others conduct human
    evaluation \cite{Krause:2017a,Fang:2018}, and some use both
    \cite{Shen:2018a,Xu:2018a,Baheti:2018,Ram:2018}.
    
    We conduct an extensive automatic evaluation using our \textsc{dialog-eval}
    repository\footnote{\url{https://github.com/ricsinaruto/dialog-eval}}, which
    implements 17 metrics used frequently in the literature. These are described in detail by our previous study on metrics \cite{Csaky:2019}.
    The metrics assess individual response quality, dialogue-level
    evaluation is left for future work\footnote{We believe that
    Gutenberg would perform especially well in dialogue-level metrics,
    since it contains high-quality extracted dialogues compared to the
    non-segmented noisy Opensubtitles utterances.}.  In all tables that follow,
    metrics are listed in the following order:
    response length (\(|U|\)), i.e. average number of
    words in a response. Per-word and per-utterance unigram (\(H_{w}^{u}\),
    \(H_{u}^{u}\)) and bigram (\(H_{w}^{b}\), \(H_{u}^{b}\)) entropy, measuring
    the non-genericness of responses \cite{Serban:2017b}. Unigram and
    bigram-level KL divergence (\(D_{kl}^{u}\), \(D_{kl}^{b}\)) between model
    and ground truth response sets \cite{Csaky:2019}. Embedding metrics
    \textit{average} (\textsc{avg}), \textit{extrema} (\textsc{ext}), and
    \textit{greedy} (\textsc{gre}) measuring similarity between response and
    target embeddings \cite{Liu:2016}. Coherence (\textsc{coh}), the
    cosine similarity between pairs of input and response \cite{Xu:2018}.
    Distinct-1 and distinct-2 (d1, d2) measuring the ratio of unique
    unigrams/bigrams in all responses \cite{Li:2016d}. The 4 BLEU metrics (b1,
    b2, b3, b4), measuring overlaps between respective n-grams (n=1,2,3,4) of
    response and target \cite{Shen:2018a,Xu:2018}. As discussed in \citet{Csaky:2019}, these metrics have been selected to provide a diverse evaluation measuring various aspects of response quality. Generally, we should assess response quality jointly as looking at individual metrics can be misleading.

    \begin{table*}[ht!]
        \begin{subtable}[t]{0.99\textwidth}
            \centering
            \small
            \renewcommand{\arraystretch}{1.0}
            \begin{tabular}[t]{p{0.1cm}p{0.1cm}p{0.1cm}R{0.6cm}p{0.4cm}p{0.3cm}R{0.5cm}R{0.5cm}R{0.6cm}R{0.6cm}p{0.4cm}p{0.4cm}p{0.4cm}p{0.4cm}p{0.5cm}p{0.4cm}p{0.3cm}p{0.3cm}p{0.3cm}p{0.3cm}}
                
                &&& \(|U|\) & \(H_{w}^{u}\) & \(H_{w}^{b}\) & \(H_{u}^{u}\) & \(H_{u}^{b}\) & \(D_{kl}^{u}\) & \(D_{kl}^{b}\) & \textsc{avg} & \textsc{ext} & \textsc{gre} & \textsc{coh} & d1 &d2&b1&b2&b3&b4\\ \midrule
                
                \multirow{5}{*}{\bf \rotatebox{90}{Transformer}}&
                \multirow{2}{*}{\bf \rotatebox{90}{ZS}} &\textsc{g}&\bf7.5&\bf6.92&\bf11.7&\bf52&\bf71&\bf.90&\bf1.72&\bf.522&\bf.509&\bf.577&\bf.579&\bf.0251&\bf.110&\bf.098&\bf.095&\bf.091&\bf.083\\
                &&\textsc{o}&4.8&6.65&10.6&32&41&2.00&3.58&.461&.481&.533&.458&.0009&.002&.075&.068&.063&.056\\ \cmidrule{3-20}

                &\multirow{3}{*}{\bf \rotatebox{90}{FT}} 
                &\textsc{g}&8.7&\bf7.09&\bf11.8&62&87&\bf.51&\bf1.03&\bf.551&\bf.535&\bf.598&\bf.580&\bf.0292&\bf.147&\bf.140&\bf.132&\bf.126&\bf.115\\
                &&\textsc{o}&8.8&6.68&10.2&59&80&2.93&4.15&.486&.477&.560&.482&.0020&.005&.106&.117&.118&.110\\ 
                &&\textsc{b}&\bf9.9&\bf7.11&11.5&\bf71&\bf94&.88&1.60&.519&.514&.579&.525&.0132&.063&.127&.128&\bf.127&\bf.117\\ \midrule
                
                \multirow{5}{*}{\bf \rotatebox{90}{GPT2}}&
                \multirow{2}{*}{\bf \rotatebox{90}{ZS}} &\textsc{g}&\bf9.1&\bf7.53&\bf12.7&\bf70&\bf98&\bf.30&\bf.71&\bf.538&\bf.500&\bf.564&\bf.559&.0333&.226&\bf.104&\bf.109&\bf.108&\bf.101\\
                &&\textsc{o}&5.7&7.19&12.3&42&56&.32&.81&.491&.484&.554&.532&\bf.0463&\bf.249&.082&.079&.076&.069\\  \cmidrule{3-20}

                &\multirow{3}{*}{\bf \rotatebox{90}{FT}} 
                &\textsc{g}&9.6&7.61&12.7&75&105&.12&\bf.33&\bf.568&\bf.540&\bf.596&\bf.573&.0407&.259&\bf.151&\bf.143&\bf.139&\bf.128\\
                &&\textsc{o}&9.4&7.62&12.6&74&102&.14&.40&.561&.533&.589&\bf.574&.0455&.264&.142&.136&.132&.122\\ 
                &&\textsc{b}&\bf10.0&\bf7.76&\bf12.8&\bf80&\bf109&\bf.11&.36&\bf.567&.535&.589&\bf.576&\bf.0486&\bf.285&.147&\bf.143&\bf.141&\bf.130\\ \midrule
                
                \multicolumn{3}{c}{\bf RT}
                &13.6&8.41&14.1&118&179&.03&.17&.496&.461&.523&.493&.0693&.414&.086&.117&.127&.122\\
                \multicolumn{3}{c}{\bf GT}
                &13.8&8.38&13.7&117&152&0&0&1&1&1&.572&.0587&.400&1&1&1&1\\
                
            \end{tabular}
            \caption{DailyDialog test set}
            \label{table:transfer_a}
        \end{subtable}
        
        \bigskip
        
        \begin{subtable}[t]{0.99\textwidth}
            \centering
            \small
            \renewcommand{\arraystretch}{1.0}
            \begin{tabular}[t]{p{0.1cm}p{0.1cm}p{0.1cm}R{0.6cm}p{0.4cm}p{0.4cm}p{0.4cm}R{0.5cm}R{0.6cm}R{0.6cm}p{0.4cm}p{0.4cm}p{0.4cm}p{0.4cm}p{0.5cm}p{0.4cm}p{0.3cm}p{0.3cm}p{0.3cm}p{0.3cm}}
                
                &&& \(|U|\) & \(H_{w}^{u}\) & \(H_{w}^{b}\) & \(H_{u}^{u}\) & \(H_{u}^{b}\) & \(D_{kl}^{u}\) & \(D_{kl}^{b}\) & \textsc{avg} & \textsc{ext} & \textsc{gre} & \textsc{coh} & d1 &d2&b1&b2&b3&b4\\ \midrule
                
                \multirow{5}{*}{\bf \rotatebox{90}{Transformer}}&
                \multirow{2}{*}{\bf \rotatebox{90}{ZS}} &\textsc{g}&\bf8.3&\bf6.99&\bf11.9&\bf57.7&\bf80&\bf1.00&\bf2.24&\bf.493&.540&\bf.545&\bf.574&\bf.0154&\bf.077&.091&.092&.091&.084\\
                &&\textsc{o}&6.6&6.70&11.5&45.2&67&2.00&2.85&.471&\bf.556&.542&.476&.0004&.001&\bf.094&\bf.098&\bf.095&\bf.088\\ \cmidrule{3-20}

                &\multirow{3}{*}{\bf \rotatebox{90}{FT}} 
                &\textsc{g}&\bf11.0&6.48&10.4&68.2&92&\bf1.28&\bf2.15&\bf.513&\bf.575&\bf.571&\bf.593&\bf.0104&\bf.048&\bf.165&\bf.163&\bf.164&\bf.155\\
                &&\textsc{o}&10.6&6.37&10.1&68.3&98&2.58&2.66&.431&\bf.575&.532&.444&.0011&.002&.148&.151&.154&.146\\ 
                &&\textsc{b}&\bf11.1&\bf6.88&\bf11.0&\bf76.5&\bf110&\bf1.28&2.21&.508&.570&.562&.559&.0047&.018&\bf.164&\bf.163&\bf.165&\bf.156\\\midrule
                
                \multirow{5}{*}{\bf \rotatebox{90}{GPT2}}&
                \multirow{2}{*}{\bf \rotatebox{90}{ZS}} &\textsc{g}&\bf9.5&\bf7.62&\bf13.1&\bf72.7&\bf101&.56&1.15&\bf.510&\bf.501&\bf.531&\bf.551&.0206&.160&\bf.092&\bf.104&\bf.107&\bf.101\\
                &&\textsc{o}&6.0&7.35&12.6&44.9&60&\bf.44&\bf1.11&.478&.491&.519&.537&\bf.0294&\bf.186&.072&.074&.072&.066\\ \cmidrule{3-20}

                &\multirow{3}{*}{\bf \rotatebox{90}{FT}} 
                &\textsc{g}&\bf11.0&7.45&\bf11.8&\bf82.6&\bf116&.27&.64&\bf.536&\bf.559&\bf.558&\bf.590&.0182&.129&\bf.157&\bf.159&\bf.162&\bf.153\\
                &&\textsc{o}&10.5&7.41&11.6&78.1&108&.32&.71&.531&\bf.558&.555&.583&.0205&.129&.153&.154&.155&.146\\ 
                &&\textsc{b}&10.3&\bf7.50&\bf11.8&77.9&108&\bf.25&\bf.61&.533&.554&.553&.587&\bf.0219&\bf.136&.151&.154&.155&.146\\\midrule

                \multicolumn{3}{c}{\bf RT} &11.6&8.51&14.0&98.6&148&.03&.14&.489&.499&.496&.488&.0495&.350&.099&.127&.136&.131\\
                \multicolumn{3}{c}{\bf GT}&11.5&8.46&13.4&97.3&124&0&0&1&1&1&.559&.0421&.337&1&1&1&1\\
                
            \end{tabular}
            \caption{PersonaChat test set}
            \label{table:transfer_b}
        \end{subtable}
        \caption{
            Metrics computed on the test set of DailyDialog and PersonaChat for Transformer and GPT2 trainings.  Pre-trained models on
            Gutenberg (\textsc{g}) and Opensubtitles (\textsc{o}) are compared.
            \textsc{b} is a Transformer or GPT2 baseline trained only on the
            small datasets, evaluated at the validation loss minimum.
            \textsc{rt} refers to randomly selected responses from the
            DailyDialog or PersonaChat training set, and \textsc{gt} to the
            ground truth response set.  Best results (with a 95\% confidence
            interval) are highlighted separately for the zero-shot
            (\textsc{zs}) and finetuned (\textsc{ft}) scenarios.
        }
        
        \label{table:transfer}
    \end{table*}

    \begin{table*}[h!]
        \begin{center}
            \small
            \renewcommand{\arraystretch}{1.0}
            \begin{tabular}{p{4.5cm}p{1.5cm}p{2cm}p{1.5cm}p{1.5cm}p{2cm}}
                \bf Input & \bf TRF & \bf GPT2 & \bf GUT ZS & \bf OPEN FT & \bf GUT FT \\ \midrule
                
                how are you doing today  EOU  awesome . just sitting here listening to some stones . how are you ?  EOU  i 'm good . just about to play some dd & what do you play ? i 'm a professional athlete . & what kind of music do you play ? & huh ! what do you think of that ? & i 'm just thinking about having a ball . & what are you watching ?\\ \midrule
                lol i scratch my nose when oreo shakes tickle it  EOU  you are funny . not so depressed about welfare and dad in ace rated now  EOU  ouch sorry your dad in jail mine passed before he could teach me to drive & i 'm so sorry . i 'm so sorry . & lol you should take him to the doctor . it is a good therapy & but you 're not a boy any more . & i think he was . my mom was not there either & oh okay . he 's a detective . my dad passed before he could train me\\ \midrule
                i used to drive my mother crazy i liked to smoke i am tee total now though  EOU  i just finish smoking . i take anxiety medication at night  EOU  that 's not good having anxiety i mean . does the medication help ? & i do not know . i am very sorry . & it works for me but i have to get up early so i can smoke & you 're perfectly right you ca n't do much harm . & it does help . i am a teacher . i teach middle school kids & no but i can be on the watch with my friends . i 'm getting to see my dad\\ \midrule
                i 've one dog she 's amazing .  EOU  i 've 2 dogs . i should take them walking instead of eating .  EOU  that would be a great idea . & i 'm so excited to go to the movies . & i think that 's too bad for my health . & very well . it 's a bargain . & what do you do for a living ? & i 've no opinion about it . i 'm very passionate about animals .\\ \midrule
                wonderful . i hail from germany . what about you ? & i 'm a small world . i 'm a waitress . & i 'm from chicago how about you ? & i 'm a british spy . & i am a student in the us . & i 'm from baltimore and i 'm also from florida\\

            \end{tabular}
        \end{center}
        \caption{\label{table:personachat_samples} Random test samples from
        PersonaChat. \textsc{trf} is the base Transformer and \textsc{gpt2} is the
        non-pretrained GPT2 model. \textsc{gut} and \textsc{open} refer to
        Gutenberg and Opensubtitles, respectively, and \textsc{zs} and \textsc{ft}
        refer to zero-shot and finetuned settings, respectively. EOU means ``End Of Utterance''.
        }
    \end{table*}

    \begin{table*}[ht!]
        \begin{center}
            \small
            \renewcommand{\arraystretch}{0.9}
            \begin{tabular}{p{0.1cm}p{0.2cm}R{0.6cm}p{0.4cm}p{0.3cm}R{0.6cm}R{0.6cm}R{0.6cm}R{0.6cm}p{0.3cm}p{0.3cm}p{0.3cm}p{0.4cm}p{0.4cm}p{0.4cm}p{0.5cm}p{0.4cm}p{0.5cm}p{0.4cm}}
                
                && \(|U|\) & \(H_{w}^{u}\) & \(H_{w}^{b}\) & \(H_{u}^{u}\) & \(H_{u}^{b}\) & \(D_{kl}^{u}\) & \(D_{kl}^{b}\) & \textsc{avg} & \textsc{ext} & \textsc{gre} & \textsc{coh} & d1 &d2&b1&b2&b3&b4\\ \midrule
                
                \multirow{4}{*}{\bf \rotatebox{90}{EN}} &\textsc{g}&\bf8.8&\bf7.77&13.4&\bf69&\bf105&.331&.707&\bf.494&.468&.518&\bf.529&.0034&.037&.0806&\bf.0879&\bf.0883&\bf.0828\\
                &\textsc{o}&6.1&7.68&13.4&47&68&\bf.292&\bf.689&.472&\bf.475&\bf.522&.519&\bf.0048&\bf.045&\bf.0867&.0855&.0810&.0739\\
                &\textsc{rt}&14.3&9.21&16.4&135&223&.038&.148&.462&.443&.485&.462&.0139&.150&.0671&.0879&.0946&.0915\\
                &\textsc{gt}&14.1&9.14&16.0&132&208&0&0&1&1&1&.526&.0089&.130&1&1&1&1\\ \midrule
                
                \multirow{4}{*}{\bf \rotatebox{90}{DE}} &\textsc{g}&\bf7.4&7.98&13.9&\bf60&\bf84&\bf.194&\bf.500&\bf.536&.581&\bf.581&\bf.576&\bf.0387&\bf.241&.0803&.0813&.079&.0734\\
                &\textsc{o}&6.4&\bf8.12&\bf14.3&52&72&.269&.635&.524&.581&.579&.566&.0329&.236&\bf.0825&\bf.0864&\bf.083&\bf.0769\\
                &\textsc{rt}&15.6&9.47&16.5&152&246&.106&.265&.519&.548&.560&.518&.0910&.453&.0723&.0946&.101&.0984\\
                &\textsc{gt}&15.0&9.15&15.5&139&186&0&0&1&1&1&.583&.0610&.392&1&1&1&1\\ \midrule
                
                \multirow{4}{*}{\bf \rotatebox{90}{NL}} &\textsc{g}&\bf6.8&7.81&13.8&\bf53&\bf76&\bf.214&\bf.624&.503&.526&.581&.541&\bf.0453&\bf.282&\bf.0858&.0854&.083&.077\\
                &\textsc{o}&5.8&7.79&\bf14.0&45&64&.388&.922&.504&.524&.580&.543&.0382&.252&.0850&\bf.0869&\bf.084&.077\\
                &\textsc{rt}&15.4&9.15&16.0&143&233&.155&.455&.513&.505&.566&.512&.0961&.487&.0855&.108&.115&.111\\
                &\textsc{gt}&14.4&9.04&15.5&129&172&0&0&1&1&1&.558&.0659&.404&1&1&1&1\\ \midrule
                
                \multirow{4}{*}{\bf \rotatebox{90}{ES}} &\textsc{g}&\bf8.0&7.16&12.1&\bf58&\bf83&.373&.744&\bf.452&.471&\bf.524&.473&.056&.242&\bf.0883&\bf.0839&\bf.0788&\bf.0723\\
                &\textsc{o}&5.8&\bf7.76&\bf13.4&46&61&\bf.198&\bf.621&.438&.466&.516&\bf.507&\bf.093&\bf.397&.0840&.0771&.0716&.0642\\
                &\textsc{rt}&12.2&8.95&15.3&111&174&.127&.226&.429&.421&.495&.426&.180&.633&.0763&.0908&.0936&.0896\\
                &\textsc{gt}&14.5&8.47&14.1&122&153&0&0&1&1&1&.490&.119&.502&1&1&1&1\\ \midrule
                
                \multirow{4}{*}{\bf \rotatebox{90}{IT}} &\textsc{g}&\bf6.9&7.59&12.7&\bf51&\bf69&\bf.183&\bf.331&\bf.452&.486&.544&\bf.490&.131&.451&\bf.0732&\bf.0746&\bf.0708&\bf.0658\\
                &\textsc{o}&4.9&\bf7.89&\bf13.6&39&49&.266&.987&.434&.485&.538&.473&\bf.155&\bf.558&.0676&.0638&.0604&.0551\\
                &\textsc{rt}&12.7&9.24&15.5&119&182&.163&.280&.452&.452&.518&.453&.253&.755&.0668&.0801&.0827&.0797\\
                &\textsc{gt}&14.6&8.64&14.0&123&138&0&0&1&1&1&.522&.182&.614&1&1&1&1\\ \midrule
                
                \multirow{4}{*}{\bf \rotatebox{90}{HU}} &\textsc{g}&4.59&7.62&\bf13.2&34.3&38&\bf.176&.530&\bf.410&.452&.520&.447&\bf.120&\bf.463&.086&.075&.0677&.0609\\
                &\textsc{o}&\bf5.56&\bf7.73&13.0&\bf42.1&\bf44&.278&.538&.401&.447&\bf.529&.442&.111&.419&\bf.106&\bf.100&\bf.0937&\bf.0848\\
                &\textsc{rt}&9.62&9.68&15.6&95.5&136&.195&.355&.393&.406&.487&.391&.305&.788&.075&.087&.0893&.0849\\
                &\textsc{gt}&7.71&9.04&14.8&65.5&72&0&0&1&1&1&.440&.220&.658&1&1&1&1\\ \midrule
                
                \multirow{4}{*}{\bf \rotatebox{90}{PT}} &\textsc{g}&\bf8.4&7.44&12.6&\bf63&\bf88&.189&\bf.495&.455&.409&.552&.474&.184&.575&.0886&.0933&.093&\bf.087\\
                &\textsc{o}&6.3&7.62&\bf13.0&49&61&.226&.671&.443&.407&.544&.488&\bf.210&\bf.627&.0816&.0812&.078&.072\\
                &\textsc{rt}&14.5&9.16&15.2&134&207&.118&.415&.441&.368&.503&.441&.316&.821&.0784&.0971&.104&.100\\
                &\textsc{gt}&17.1&9.02&14.8&156&235&0&0&1&1&1&.506&.249&.712&1&1&1&1\\

            \end{tabular}
        \end{center}
        \caption{\label{table:languages_new} Comparing Gutenberg and Opensubtitles
        GPT2 trainings across 7 languages on the union of the two test sets.
        The second column shows whether the model was trained on Gutenberg
        (\textsc{g}) or Opensubtitles (\textsc{o}). Randomly selected responses
        from the respective train set (\textsc{rt}) and groud truth
        (\textsc{gt}) performance is also given. Significantly better results
        between Gutenberg and Opensubtitles (95\% confidence interval) are
        highlighted on each test set.
        }
    \end{table*}

    \subsection{Trainings}
    
    We conduct experiments with
    Transformer\footnote{\url{https://github.com/tensorflow/tensor2tensor}} and
    GPT2\footnote{\url{https://github.com/huggingface/transfer-learning-conv-ai}}
    models. The Transformer is trained on utterance pairs, and we use the base
    version of roughly 50M parameters (further training details are given in \Cref{ssec:hyper_appendix}). The
    vocabulary is set to the top 100 000 words for Gutenberg and Opensubtitles
    trainings, and 32 768 and 16 384, for PersonaChat and DailyDialog,
    respectively. The Transformer is trained for 21 epochs on Gutenberg and
    Opensubtitles, because of time and hardware constraints, but the validation
    loss was still decreasing. Training took about 80 hours on a single RTX
    2080 Ti, with batch size set to the memory limit. We used the Adam optimizer
    \cite{Kingma:2014}. For generating test outputs greedy decoding is used.
    
    For the GPT2 trainings (117M pretrained version) we set the maximum number
    of previous utterances to be used as history to 3 (parameter details in \Cref{ssec:hyper_appendix}). The huggingface repository leverages GPT2 for dialogue modeling
    with an additional personality input and a random candidate classification
    loss \cite{Wolf:2019}. We set the personality field to empty and use a
    single random candidate response from the training set for each example. We
    use the nucleus sampling implementation in the repository with default
    parameters to sample outputs \cite{Holtzman:2020}. All GPT2 trainings are trained with a batch size of 2 and evaluated at the minimum of the validation loss. The English GPT2 Gutenberg
    training took about 20 days (7 epochs), on an RTX 2080 Ti, while on
    Opensubtitles the validation minimum was reached after a single epoch of
    training (about 2 days). Finetuning on DailyDialog and PersonaChat and
    trainings on other languages took generally less than 1 day, except the
    German trainings (2 days).
    
    We evaluate Gutenberg and Opensubtitles pre-trained models in zero-shot and
    finetuning scenarios on DailyDialog and PersonaChat.  The same amount of
    training data and train/test/dev ratio is used for both Gutenberg and
    Opensubtitles. Models are finetuned until the validation
    loss minimum is reached. Finetuning experiments are only done
    in English, due to the lack of additional datasets in other languages.
    For Transformer
    trainings, we remove overlapping utterance pairs between the official train
    and test sets from the DailyDialog training set. We observed that inflated
    results reported on DailyDialog \cite{Csaky:2019} are partly due to this
    overlap. For all datasets we use lowercase input text and \textsc{nltk}\footnote{\url{https://www.nltk.org/}} word
    tokenization as preprocessing. We use the official DailyDialog splits and we employ a random train/dev/test split of 80/10/10 for PersonaChat, which we make publicly available along all the datasets used in this paper\footnote{\url{https://github.com/ricsinaruto/gutenberg-dialog}}.

    Gutenberg pre-training performs better than Opensubtitles on DailyDialog
    across nearly all metrics in both zero-shot and finetuned settings
    (\Cref{table:transfer_a}). Gutenberg pre-training outperforms even the
    model trained only on DailyDialog on some metrics. 
    All GPT2 models are pretrained as language models on web text. Thus it
    comes as no surprise that the additional pretraining on Gutenberg does not
    lead to the same relative improvement as with the Transformer models, which
    are trained from scratch. Gutenberg pre-training achieves better results
    than Opensubtitles in all metrics after finetuning on PersonaChat
    (\Cref{table:transfer_b}). In the Transformer zero-shot scenario,
    Opensubtitles achieves better BLEU scores, however, zero-shot BLEU scores
    are generally much lower than randomly selected responses, questioning the
    validity of this comparison. Gutenberg pre-training outperforms the
    baseline PersonaChat training on some metrics after finetuning. Considering
    the domain mismatch between the older Gutenberg books and the modern
    chit-chat style datasets this is especially impressive. Since the metrics
    are all very similar it is also important to look at responses
    qualitatively. \Cref{table:personachat_samples} presents 5 random test
    samples. 
    More samples from both DailyDialog and PersonaChat can be found in \Cref{ssec:responses_appendix}. It is clear that the Transformer and the zero-shot GPT2 scenario
    perform the worst, followed by the finetuned Opensubtitles training. This
    shows some anecdotal support for the effectiveness of pre-training on
    Gutenberg.
    
    \Cref{table:languages_new} compares Gutenberg and Opensubtitles trainings
    across all seven languages, using roughly the same amount of data. In
    absence of a third independent data source we create mixed test
    datasets for each language that include the same amount of data from
    Gutenberg and Opensubtitles, by limiting the larger of the two to the size
    of the smaller.
    Except for Hungarian, models trained on Gutenberg perform better on more metrics than Opensubtitles trainings. On some metrics, models perform
    worse than random responses from the training set. This is expected for
    entropy and distinct metrics, but we believe that BLEU scores would be higher
    after further training since overfitted models have been shown to perform better on these metrics \cite{Csaky:2019}.
    This lack of stopping criteria also
    makes a fair comparison challenging. Example responses from all models are
    shown in \Cref{ssec:responses_appendix}. To our knowledge, this is the first work to use
    non-English languages from the Opensubtitles dataset for dialogue modeling,
    and there are very few chatbot models in non-English languages in general.

    \section{Conclusion}
    \label{sec:conclusion}
    We presented the Gutenberg Dialogue Dataset consisting of 14.8M utterances
    in English and smaller datasets in German, Dutch, Spanish, Italian,
    Hungarian, and Portuguese. We described heuristics used in our dialogue
    extraction pipeline and conducted a detailed error analysis to uncover the
    causes of errors and to assess data quality. In a pre-training comparison
    between Gutenberg and Opensubtitles we found that Gutenberg performs better
    on downstream datasets 
    in both zero-shot and finetuning scenarios. We release the Gutenberg dataset as well as the
    open-source pipeline\footnote{We also release all data, trained models, and training scripts to produces the results.} with which researchers can build their own datasets.  We also built a web demo interface to all models presented in the paper\footnote{\url{https://ricsinaruto.github.io/chatbot.html}}.
    
    In future work, we wish to improve heuristics and dataset quality. A
    classifier could be trained to decide whether two consecutive utterances
    are part of the same dialogue (looking at non-conversational context). Positive and negative examples could be generated by a very
    low/high dialogue gap, or by manual annotation. Speaker-related errors
    could be addressed
    using speaker identification. We also hope to extend our
    dataset to more languages. This involves delimitation
    analysis, implementation of heuristics, and error analysis. We welcome
    contributions from the community, as our open-source modular pipeline
    minimizes the effort required for adding new languages.

    \section*{Acknowledgments}
    We wish to thank M\'{a}rton Makrai for inspiration and discussion about the idea of extracting dialogue from books. We wish to thank members of the SZTAKI HLT\footnote{\url{https://hlt.bme.hu/en/}} group and all anonymous reviewers for their help and thoughtful feedback. Work partly supported by Project FIEK 16-1-2016-0007, financed by the FIEK 16 funding scheme of the Hungarian National Research, Development and Innovation Office (NKFIH). Recski was partly supported by BRISE-Vienna (UIA04-081), a European Union Urban Innovative Actions project.
    
        \bibliography{ml}
    \bibliographystyle{acl_natbib}
    
    \clearpage
    \appendix
    \section{Appendix}
    
	\subsection{Training hyperparameters}
	\label{ssec:hyper_appendix}
	\begin{table}[h!]
		\begin{center}
			\begin{tabular}{lr}
			\toprule
				\bf Name & \bf Value \\ \midrule
				
				Hidden size & 512 \\
				Number of hidden layers & 6 \\
				Label smoothing & 0.1 \\
				Filter size & 2048 \\
				Number of attention heads & 8 \\
				Layer dropout & 0.1 \\
				Relu dropout & 0 \\
				Attention dropout & 0 \\
				Learning rate & 0.2 \\
				Learning rate warmup steps & 8000 \\
				\bottomrule
			\end{tabular}
		\end{center}
		\caption{Transformer hyperparameters.}
	\end{table}
	
		\begin{table}[h!]
		\begin{center}
			\begin{tabular}{lr}
			\toprule
				\bf Name & \bf Value \\ \midrule
				
				 LM loss coefficient & 2 \\
				 Multiple-choice loss coefficient & 1 \\
				 Max. gradient norm & 1 \\
				 Gradient accumulation steps & 8 \\
				 Nucleus sampling p & 0.9 \\
				 Context size & 1024 \\
				 Embedding size & 768 \\
				 Number of attention heads & 12 \\
				 Number of layers & 12 \\
				 Vocabulary size & 50262 \\
				\bottomrule
			\end{tabular}
		\end{center}
		\caption{GPT2 hyperparameters.}
	\end{table}
	
    \subsection{Gutenberg statistics}
    \label{ssec:stat_appendix}
    
    \begin{figure}[!ht]
        \centering
        \includegraphics[width=0.48\textwidth]{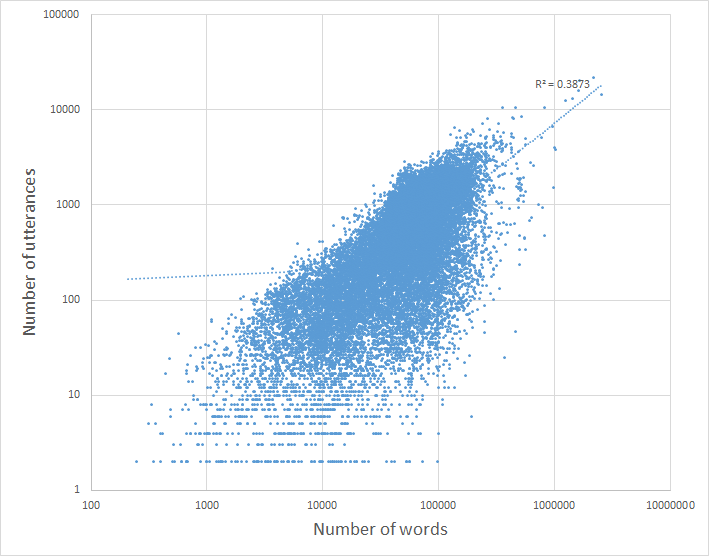}
        \caption{Number of extracted utterances with respect to number of words in each book on logarithmic scales (English Gutenberg dataset).}
        \label{fig:delimiters1}
    \end{figure}
    
    \begin{figure*}[!htb]
        \centering
        \includegraphics[width=0.9\textwidth]{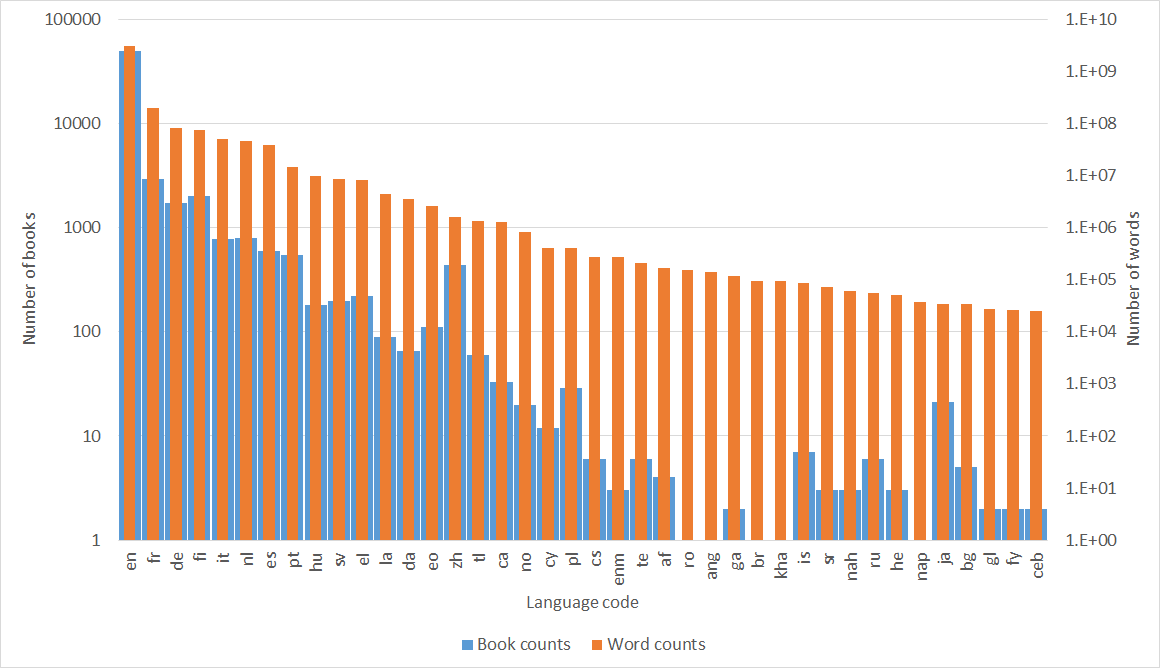}
        \caption{Number of books and words for top 40 languages in Project Gutenberg on logarithmic scales.}
        \label{fig:delimiters2}
    \end{figure*}

    \begin{figure*}[!htb]
        \centering
        \includegraphics[width=0.9\textwidth]{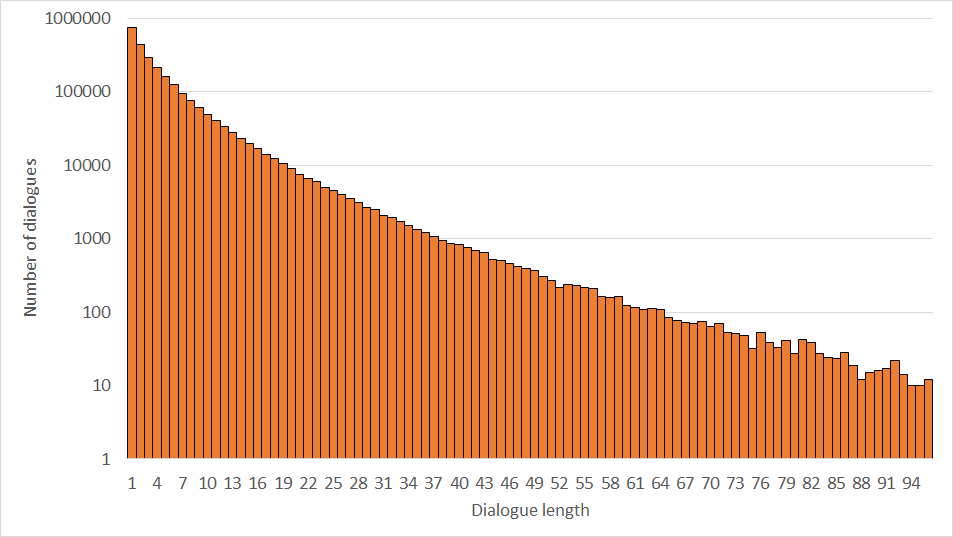}
        \caption{Distribution of dialogue lengths in the English Gutenberg dataset (truncated).}
        \label{fig:delimiters3}
    \end{figure*}

	\subsection{Example responses}
	\label{ssec:responses_appendix}
	
	\begin{table*}[h!]
		\begin{center}
			\small
			\renewcommand{\arraystretch}{1.1}
			\begin{tabular}{p{5cm}p{1.5cm}p{1.5cm}p{1.5cm}p{2cm}p{2cm}}
			    \toprule
				\bf Input & \bf TRF & \bf GPT2 & \bf GUT ZS & \bf OPEN FT & \bf GUT FT \\ \midrule
				
				stay for dinner with us .  EOU  i 'm afraid i ca n't . i have a dinner party at seven tonight . you see i am going to england the day after tomorrow .  EOU  well i wo n't keep you then . it was very nice of you to come to see me . & you 're welcome . & thank you . i will . & i knew you would n't mind . & yes . i hope you feel better . & it was my pleasure .\\\midrule
				sure . take a seat . . . what would you like to know ?  EOU  first i ve heard that your latest book is based on a true story .  EOU  that s correct . it s a murder mystery based on actual murders that book place in florida several years ago . the main characterthe police invesigatoris based on the man who investigated the case . & i see . i thought it was a great idea . i could do that . & oh i see . thanks a lot . i ll have to check it out . & well ? & i ve never been involved in an actual murder . what are your chances of catching the killer ? & so what happened to the man who investigated the case ?\\\midrule
				well i do n't eat any meat but i do eat fish and eggs .  EOU  do you eat a lot of vegetables ?  EOU  oh yes and fruit i love fruit . & what kind of food do you like ? & do you like fruits and vegetables ? & and do you eat a good many things ? & do you like fruits ? & what do you do with your vegetables ?\\\midrule
				that 's terrific . i ca n't wait to check it out .  EOU  hang on . are you sure you want to keep a nontraditional pet ?  EOU  why not ? it 's so cool . & i do n't know . i do n't know what to do . & the pets come with a 2 week guarantee . & i 've been in the market for two years . & i 'm really glad that you came . & then let 's go .\\\midrule
				that 's my boy ! katherine curtis created this activity in 1920 . and in 1984 it was authorized as one activity in olympic games .  EOU  wow mom is more knowledgeable than dad . i must learn from you .  EOU  my little boy you should learn more from me . i 'm almost thirty years older than you . & i 'm not sure . i 'm a bit nervous . & dad i 'm not a child anymore . i 'm a grownup . & you 're a nice old lady mom . & okay . i will begin learning when i grow up . & what 's the matter with your boy ?\\ \midrule
				it will cost you 10 .  EOU  by the way what 's the time ?  EOU  you can see the time on the meter . i have radio time you know . & i know . i do n't know what to do . & that 's fine . i 'll see it tomorrow . & you had a good time in new york ? & you can watch the news . & i see .\\\midrule
                hello . is bob there please ? & hello . this is mr . green . & yes he is . how 's your business going ? & yes my dear but is n't he coming ? & yes . he 's gone out to the market . & speaking .\\\midrule
                it seldom rains this summer . & i think it 's a bit expensive . & why ? & you are always on the watch for rain & i am sure it does . & what is the weather like today ?\\\midrule
                i think i may have seen it on the kitchen table . & i m afraid i have a bit of a small town . & it 's great . i 've enjoyed it . & i 'm sure you did & no i didn t . i didn t see it . & what did you see ?\\\midrule
                thanks . and could i have his email just in case i ca n't get him by phone ?  EOU  sure . his cell phone is 09112223 33 . and his email is lower case t smiththat 's one word at c c w dot com dot t w .  EOU  thank you so much . does he read his emails daily ? & yes . he was in the middle of the city . & not every day . he just sends email to his friends . & sure . he has n't been here since the night of the murder . & yes he does . he reads a lot of letters per day . & yes he does . he 's a very good student . i can read and write and he keeps in touch with my friends .\\

				\bottomrule
				
			\end{tabular}
		\end{center}
		\caption{Random test samples from DailyDialog. \textsc{trf} is the base Transformer and \textsc{gpt2} is the non-pretrained GPT2 model. \textsc{gut} and \textsc{open} refer to Gutenberg and Opensubtitles, respectively, and \textsc{zs} and \textsc{ft} refer to zeroshot and finetuned settings, respectively. EOU means ``End Of Utterance''.}
	\end{table*}
	
	\begin{table*}[h!]
		\begin{center}
			\small
			\renewcommand{\arraystretch}{1.1}
			\begin{tabular}{p{5cm}p{1.5cm}p{1.5cm}p{1.5cm}p{2cm}p{2cm}}
			\toprule
				\bf Input & \bf TRF & \bf GPT2 & \bf GUT ZS & \bf OPEN FT & \bf GUT FT \\ \midrule
				
				i do n't but that would be great  EOU  what do you do for a living ?  EOU  i work w french fries . i stink of them & that 's cool . i 'm a waitress . & do you have any pets ? & you mean to say that you do n't learn much ? & that sounds like a hard job . do you like to read ? & oh i see . i am a legal assistant .\\\midrule
great ! just came back from a walk with my bipolar dog .  EOU  that sounds nice i 've been taking care of a buddy horses this afternoon .  EOU  why ? wheres the hubby ? & i 'm not sure . i 'm not sure . & he has a dog i 've a dog too ! & he 's in the stable . & he 's a shepherd and is very smart . & i 'm trying to get him to see me .\\\midrule
that is so sad i 'm sorry  EOU  i never see my dad now he moved to guam  EOU  i do not even know where guam is & i 'm a girl . i love to eat out . & where do you live now ? & i thought he was in the penitentiary & i do n't even know if i 've ever heard of it & i have always wanted to move there but he passed away\\\midrule
that 's very interesting . i workout for a living . & i do too . i am a waitress . & i 'm retired and working with animals . & you do eh ? & that 's cool . i like to eat breakfast . & oh my god i hate eating too much . i could n't do it .\\\midrule
my one child wanted to be like that too  EOU  it is good for the body  EOU  do you train yourself alone ? & i do . i am a waitress . & i do i try to at least once a month & yes . & no i do n't have a job . & yes i have a job in a factory\\\midrule
i am 35 and still single .  EOU  are you looking forward to turning 40 ?  EOU  no not at all . it seems dreadful . & i agree . i 'm very excited about that . & i 'm sure you are . what is your name ? & i am sorry to hear it . & what do you do for work ? & i have a dog named george . what do you do for a living ?\\\midrule
being from springfield missouri the midwest does not offer much  EOU  i can not wait to grow up  EOU  do n't rush it . . . believe me stay young & i am very tall so i do not have to go to the gym & i do n't want to be the boss of a nascar car & you know what i mean i ca n't let you go . & i will try it & i will be twenty in march\\\midrule
congrats ! i have two little ones and a wife back stateside and i miss them so much !  EOU  i live with my best friend on a lake  EOU  sweet life . beats being in this desert although i got ta admit i love my long deployments & i 'm a bit of a guy . i love to swim . & i like to swim in the mountains & then why not stay at home ? & i am still in college but i am doing my masters & i 'm sorry to hear that\\\midrule
hi . how are you doing ?  EOU  well . . i got ta tell the truth to somebody i just have too  EOU  go ahead . i 'm almost seven feet tall by the way & i 'm a girl . i 'm tall . & that is tall ! you can ride horses and trucks . & oh i ai n't ! i 'm eight feet eight . & wow ! ! i 'm just like a sponge . . . i like to work with spongebob & wow that 's a huge size . i just bought a new new dress a few times .\\\midrule
hi i 'm jane and a part of my skin is blue i 'm 177 cm .  EOU  your skin is blue ? my names bill nice to meet you  EOU  yeah it was a dye i was using and it seeped into the skin . & i like to draw and paint . & oh wow that 's really pretty & then it was not a dye . & that is amazing ! what do you do for a living ? & what color was it ? i love the color red\\

				\bottomrule
				
			\end{tabular}
		\end{center}
		\caption{Random test samples from PersonaChat. \textsc{trf} is the base Transformer and \textsc{gpt2} is the non-pretrained GPT2 model. \textsc{gut} and \textsc{open} refer to Gutenberg and Opensubtitles, respectively, and \textsc{zs} and \textsc{ft} refer to zeroshot and finetuned settings, respectively. EOU means ``End Of Utterance''.}
	\end{table*}

	\begin{table*}[h!]
		\begin{center}
			\small
			\renewcommand{\arraystretch}{1.1}
			\begin{tabular}{p{6.5cm}p{2.5cm}p{2.5cm}p{2.5cm}}
			\toprule
				\bf Input & \bf Gutenberg & \bf Opensubtitles & \bf Target \\ 
				\midrule
				hogyan ? ön képes 200 forintot adni ?  EOU  nagyon .  EOU  eszerint jól áll a lapja ? & én . & nem vagyok a közelőtt . & kitűnően .\\\midrule
				hát a fejét ne kössük be ?  EOU  mivel ?  EOU  vizes kendővel . & hogyan ? & felség , azt mondja a nő egy közelést . & fölösleges .\\\midrule
				tessék kérem ! & tessék , hogy miért nem ? & tudom . . . & egyedül van ?\\\midrule
				veszünk ki más lakást .  EOU  meglövöldöz bennünket az öreg .  EOU  ne hidd azt ! & meglövöldöz bennünket , & ne légy nem akarják . & dehát miből éljünk ?\\\midrule
				barátom , jób , az eördög megszökött . a miskolczi részvényeket fölapríthatod fidibusznak .  EOU  míriz olezáncz !  EOU  barátom , jób ! szomorú hírt hozok : a török sorsjegyek elvesztették az értéküket ! & szegény asszony ! & mert őket , hogy megint mondtam . & míriz olezáncz !\\\midrule\midrule
				ott vettem ezt a kést egy kis boltban , két utcával arrébb .  EOU  hat dollárba került .  EOU  a törvény tiltja rugóskések vásárlását . & ah ! & ez egy kis kölykökre , hogy elváltoztatni az ügyvédő . & így van , megszegtem a törvényt .\\\midrule
				honnan szerezte ?  EOU  tegnap este a tárgyalás után sétálni mentem .  EOU  a fiú lakása környékén . & hát nem szerezni , hogy nagyon szerezni ! & én volt rá , mi történt ? & ott vettem ezt a kést egy kis boltban , két utcával arrébb .\\\midrule
				oké .  EOU  esetleg megegyezhetnénk egy határidőben .  EOU  még egy tiszteletkör , és kész . & egy határ úr , hogy öreg asszonyt üve & és előtt a gyilkosságot . & most negyed 7 van .\\\midrule
				ki mondja , hogy nem bűnös ?  EOU  egy .  EOU  rendben . & kicsoda ? & csak az ügyvéd volt . & tizenegy bűnös , egy nem bűnös .\\\midrule
				én egyszerűen csak kérdéseket teszek fel .  EOU  ez most jutott eszembe . . .  EOU  ha én öltem volna meg az apámat , én is visszamentem volna a késért . & ugyan , ugyan ! & a feleségét . & a srác azt hihette , hogy nem látták , amikor elrohant , és másnap reggelig nem fedezik fel \\
				
				\bottomrule

			\end{tabular}
		\end{center}
		\caption{Random test samples from the Hungarian Gutenberg test set (first 5 rows), and Opensubtitles test set (last 5 rows). Responses from the GPT2 model trained on Gutenberg and Opensubtitles are compared. EOU means ``End Of Utterance''.}
	\end{table*}
	
	\begin{table*}[h!]
		\begin{center}
			\small
			\renewcommand{\arraystretch}{1.1}
			\begin{tabular}{p{6.5cm}p{2.5cm}p{2.5cm}p{2.5cm}}
			\toprule
				\bf Input & \bf Gutenberg & \bf Opensubtitles & \bf Target \\ 
				\midrule
				
				madame if i had the tact and wit of my brother charles i would say that the sun which so lately has risen must not set so soon upon its worshippers . but answer me one questionwhat is the meaning of the riddle with which your majesty has been entertaining us ?  EOU  may i answer with another question ? tell me sire have you ever seen the sun rise ?  EOU  i ? no your majesty . i confess that i never have . & well sire then i will tell you a secret . & the sun rises in the sky . & and you ladies and gentlemen ?\\\midrule
				wo n't he confess how he became possessed of the diamonds ?  EOU  no not to me . he refuses to say a word in his own defence .  EOU  then i tell you what this quixotic young man is defending another person he is shielding the assassin . & you think he is innocent ? & the longer i wait he wo n't tell me anything . & if he is that shows him to be an accessory either before or after the fact but who is the person you\\\midrule
				and who ought of right to avenge him ?  EOU  sir the knight that was in the red launde at the assembly that jousted with messire gawain and had the prize of the tournament .  EOU  did he better than messire gawain ? & he did he did indeed . & yes sir . & sir so did they adjudge him for that he was a longer time in the assembly .\\\midrule
				why do n't you keep your mouth shut philippo ? it 's always talk talk with you . we are chosen because the captain can rely upon us . & but i have no right to speak to you and i 'm not going to be treated as if i were a child . & the captain is the captain . & he can rely upon anyone who knows that he will get his throat cut if he fails in his duty .\\\midrule
				i 'd rather talk to you in my room if you please & so would i & but it 's not your fault . & all right\\\midrule\midrule
				so what 's your take on that ?  EOU  i got nothing .  EOU  what am i gon na do ? & do n't worry . i 'll keep you a prisoner . & you do n't have to do anything . & i do n't know .\\\midrule
				my face .  EOU  there was an argument .  EOU  i & what ? & i have no idea what happened . & i said some things i should n't have said and i\\\midrule
				what do you do when several people attack you at once ?  EOU  get up  EOU  no matter how many i focus on just one & one for me ! & do n't you want me to get into trouble ? & i was in a fight against five guys i went after one of them i did n't care what the other four did i\\\midrule
				i do n't know .  EOU  come with me .  EOU  come on . & i ca n't & i do n't know . & you do n't need it come on .\\\midrule
				they were all using me .  EOU  hanford everybody !  EOU  but you did n't kill everybody . & you do n't know what it means to kill people . & you 're not . & you killed randy chase .\\
				\bottomrule

			\end{tabular}
		\end{center}
		\caption{Random test samples from the English Gutenberg test set (first 5 rows), and Opensubtitles test set (last 5 rows). Responses from the GPT2 model trained on Gutenberg and Opensubtitles are compared. EOU means ``End Of Utterance''.}
	\end{table*}
	
	\begin{table*}[h!]
		\begin{center}
			\small
			\renewcommand{\arraystretch}{1.1}
			\begin{tabular}{p{6.5cm}p{2.5cm}p{2.5cm}p{2.5cm}}
			\toprule
				\bf Input & \bf Gutenberg & \bf Opensubtitles & \bf Target \\ 
				
				\midrule
				
				pues debe de haber muchas . pablo penáguilas que las sabe todas , me ha dicho que son muchas y que la vida entera de un hombre no basta para una sola . & es que no hay que dar las solas . esos que la vida no tiene nada que hacer por una sola que se les ha de hablar con el cuerpo . & no le ha dicho a nadie . & ríete tú de eso . . . . ya me verás a mí . . .\\\midrule
				¿ y por qué han de estar exentos de esa graciosa ley los solteros ricos ? ¿ por qué no han de cargar ellos también con su huérfano , como cada hijo de vecino ?  EOU  no me opongo . ¿ pero qué es esto ? . . . ¡ sangre !  EOU  ¡ jesús ! . . . . si es la nela . mira cómo se ha puesto los pies . & ¿ y no quiere usted que lo han robado ? & no . & ya se ve . . . . como tuvo que meterse entre las zarzas para coger a tu dichoso lili . nela , ven ac\\\midrule
				¿ y roberto ? & no , no me llamo . & ¿ y roberto ? & no sé , por ahí andará .\\\midrule
				echaremos por estos prados adelante . ¿ a ver qué me has traído hoy ?  EOU  busca bien y encontrarás algo  EOU  ¡ ah , madre de dios ! chocolate crudo . . . ¡ y poco que me gusta el chocolate crudo ! . . . nueces . . . una cosa envuelta en un papel . . . ¿ qué es ? ¡ ah ! ¡ madre de dios ! , un dulce . . . . ¡ dios divino ! , ¡ pues a fe que me gusta poco el dulce ! ¡ qué rico está ! en mi casa no se ven nunca estas comidas ricas , pablo . nosotros no gastamos lujo en el comer . verdad que no lo gastamos tampoco en el vestir . total , no lo gastamos en nada . & ¡ ah ! ¡ no lo sé ! & ¡ qué hizo eso ! & ¿ a dónde vamos hoy ?\\\midrule
				quita allá , tonta , tú eres una alhaja .  EOU  que no señor . si no puedo trabajar . en cuanto cargo un peso pequeño , me caigo al suelo . si me pongo a hacer alguna cosa difícil en seguida me desmayo .  EOU  todo sea por dios . . . . vamos , que si cayeras tú en manos de personas que te supieran manejar , ya trabajarías bien . & ¿ qué decís ? & ¿ se encuentra en personas ? & no , señor ; si yo no sirvo más que de estorbo .\\\midrule\midrule
				iluminó ciertos rincones oscuros del alma humana .  EOU  mostró al mundo las almas de aquellas personas que nacieron diferentes al resto .  EOU  para entender por qué estaba destinado a esto , tendríamos que retroceder en el tiempo a cualquier día de su infancia , en la ciudad de colorado springs . & ¿ y qué ? & sólo una pelea , nunca había terminado . & no tuvo una infancia fácil , ya que sus padres eran diferentes .\\\midrule
				algunos se fueron a sudamérica .  EOU  lo sé , señor .  EOU  ¿ por qué no se fue ? & ¿ por qué ? & ¿ por qué no ? & porque creo que este país no forma parte de los estados unidos .\\\midrule
				ella es muy simpática .  EOU  es mi deber .  EOU  ella es muy simpática . & pues es muy poco que una enferma . & es mi deber . & ¿ y bates ?\\\midrule
				anoche vimos un estreno en el metropolitan .  EOU  ¡ eh , es el tercero !  EOU  ¡ son tan buenos ! & ¿ son tan buenos ? & ¡ no lo se ! & ya sé que son buenos .\\\midrule
				estás equivocada .  EOU  ¿ qué más da lo que piense ?  EOU  cuando llamas a alguien asesino , le suele molestar . & ¿ y quién es el espectáculo ? & ¿ y usted ? & no se lo diré a nadie .\\
				
				\bottomrule

			\end{tabular}
		\end{center}
		\caption{Random test samples from the Spanish Gutenberg test set (first 5 rows), and Opensubtitles test set (last 5 rows). Responses from the GPT2 model trained on Gutenberg and Opensubtitles are compared. EOU means ``End Of Utterance''.}
	\end{table*}
	
	\begin{table*}[h!]
		\begin{center}
			\small
			\renewcommand{\arraystretch}{1.1}
			\begin{tabular}{p{6.5cm}p{2.5cm}p{2.5cm}p{2.5cm}}
			\toprule
				\bf Input & \bf Gutenberg & \bf Opensubtitles & \bf Target \\ 
				
				\midrule
				
				intendo di fare il mio dovere . . . denunziar tutto alla giustizia . . . la mia nomina a capo agente è combattuta dai miei rivali . . . si accorgeranno tutti sempre più che razza d'uomo è lucertolo , e che con lui non si combatte !  EOU  lucertolo ! . . .  EOU  voi non agirete così ! . . . sarebbe una grande viltà & come ? . . . & ma è che hai fatto ? & ma denunziate pure denunziate pure la ragazza di piazza degli amieri . . . come voi la chiamate . . \\\midrule
				tutti . . . chi . . . ?  EOU  il pittore . . . tu . . . la ragazza di via degli amieri . . .  EOU  la ragazza di via . . . & che cosa ? & davvero ? & eh , sì , mia cara . . . l'ho veduta un par di volte soltanto , mentre pedinavo te , e l'ho subito r\\\midrule
				due avvenimenti straordinarii si sono compiuti la sera del 14 gennaio . . . l'assassinio nel vicolo della luna . . . e un altro avvenimento al quale non ho voluto accennare nella difesa per ragioni delicate . . . & volete dire ? & e'tutto , signore . & il secondo avvenimento è . . . la sparizione di una ragazza che abitava in piazza degli amieri .\\\midrule
				ma la collana si ritroverà . . . ne sono certa - soggiungeva la principessa & che cos'ha ? & ecco . & anch'io basta guardare quella ragazza , per escludere ogni accusa come un'infamia . . .\\\midrule
				e il babbo . . . e la mamma ? & sì , il babbo . & sì , ai babbo , ai babbo . & stanno meglio , e credo che tu li potrai salvare !\\\midrule\midrule
				sono cose che non si fanno .  EOU  mi spiace , signore , ma non posso considerarla una seria obiezione .  EOU  l'ha sentito , lanyon ? & sì , & sì . & è un'altra prova della sua eccentricità ?\\\midrule
				tenete .  EOU  datele del cognac .  EOU  questo gli incoraggerà il cuore . & perdono . & non è mica una cosa . & siete una buon uomo .\\\midrule
				qual ' è il suo nome ?  EOU  parla , stupido !  EOU  mata hari . & e chi è quel che dice ? & franco , lasciami perdere . & questo è il suo nome , vero ?\\\midrule
				di cosa vivrebbe il popolo dell'altai . . . se non avesse le pecore ?  EOU  sono un'insegnante e devo parlare col capo del soviet rurale .  EOU  che cosa insegnerai ? & di che cosa c ' è ? & ci penso io . & sola . . .\\\midrule
				divertente ?  EOU  con signac ?  EOU  è necessario ? & lo spero . & si è una ragazza di peggioriose . & perchè no ?\\
				
				\bottomrule

			\end{tabular}
		\end{center}
		\caption{Random test samples from the Italian Gutenberg test set (first 5 rows), and Opensubtitles test set (last 5 rows). Responses from the GPT2 model trained on Gutenberg and Opensubtitles are compared. EOU means ``End Of Utterance''.}
	\end{table*}
	
	\begin{table*}[h!]
		\begin{center}
			\small
			\renewcommand{\arraystretch}{1.1}
			\begin{tabular}{p{6.5cm}p{2.5cm}p{2.5cm}p{2.5cm}}
			\toprule
				\bf Input & \bf Gutenberg & \bf Opensubtitles & \bf Target \\ 
				
				\midrule
				
				daarvoor is reeds gezorgd . & wat hebt gij dan te zeggen ? & kom binnen , kom binnen . & neen , o neen , naar het kot ga ik niet ! dood mij liever vr uwe voeten !\\\midrule
				men kan niet weten , 't is altijd goed , een speld bij zich te hebben .  EOU  wat wilt ge dan toch van nacht uitvoeren ?  EOU  zaken . & wat zegt ge ? & heb je ' n gezin ? & apropos .\\\midrule
				hij zegt , dat hij in elk geval binnenkomt , maar hij wou 't liever met uw toestemming doen . & maar de jongeheer van 't zelfde , h ? & waarom ben je hier ? & heb je gezegd , dat ik aan 't werk was ?\\\midrule
				wel , flipsen , wat scheelt je ? heb je je bezeerd ?  EOU  ja , burgemeester , ik heb me bezeerd ,  EOU  zoo , hoe komt dat ? & komt daar nu op den grond ! & zijn het niet ? & dat weet ik niet , burgemeester ,\\\midrule
				leve jan verhelst ! leve mie - wan na ! & leve jan verhelst ! & wat bedoel je ? & leve sander ! leve sander ! hoera ! hoera !\\\midrule\midrule
				een naald in ' n naaldberg .  EOU  en onze compagnie ?  EOU  de besten voor ons , de rest naar b . & en de rest naar b . ? & gaan jullie naar binnen ? & jezus christus .\\\midrule
				goed zo , meid .  EOU  dat is mijn molly .  EOU  gaat het goed met hem ? & wij zijn met hem , & hij is er . & ja , maar hij wil gewoon niet slapen .\\\midrule
				vijf man is ' n doel .  EOU  eentje is zonde van de munitie .  EOU  hou 't zand uit je wapen , zorg dat 't blijft werken . & vijf man is ' n doel . & je hebt haar vermoord . & tot zo , op 't strand .\\\midrule
				ik heet kovu .  EOU  ik heet kiara .  EOU  jij bent 'm . & zeg , hoeveel zijn d ' r ? & ik heet kovu . & jij bent 'm .\\\midrule
				je . . .  EOU  bent u gekomen om dat te zeggen ?  EOU  je moet naar huis . & waarom niet ? & de volgende keer niet . & we hebben bevel je terug te brengen .\\
				
				\bottomrule

			\end{tabular}
		\end{center}
		\caption{Random test samples from the Dutch Gutenberg test set (first 5 rows), and Opensubtitles test set (last 5 rows). Responses from the GPT2 model trained on Gutenberg and Opensubtitles are compared. EOU means ``End Of Utterance''.}
	\end{table*}
	
	\begin{table*}[h!]
		\begin{center}
			\small
			\renewcommand{\arraystretch}{1.1}
			\begin{tabular}{p{6.5cm}p{2.5cm}p{2.5cm}p{2.5cm}}
			\toprule
				\bf Input & \bf Gutenberg & \bf Opensubtitles & \bf Target \\ 
				
				\midrule
				
				desculpai , minha boa senhora , rosinha é minha neta .  EOU  sim , snr . a d . thereza , é minha avó , de quem tantas vezes tenho fallado a v . exc . a e . . .  EOU  então porque não continuas ? & não , snr . a d . thereza , não foi nenhuma . não sei o que é que eu digo : eu conheço - o ao sr . seabra . . . & só um bom rapaz , não . & falla , falla , minha menina . não tenhas receio . queres pedir - me alguma cousa , não é assim ?\\\midrule
				estiveste incommodada , minha filha ? & um pouco . & não , obrigado . & não , minha senhora . este cestinho , que aqui trago , é que foi a causa da minha demora .\\\midrule
				não , minha senhora . este cestinho , que aqui trago , é que foi a causa da minha demora .  EOU  como é lindo não sabia julia , que tinhas a prenda de fazer cestos de juncos entrançados .  EOU  não fui eu que fiz este cestinho , minha mãi . & e tem razão , eu não posso dizer ao senhor simão , que está a dizer que esta senhora que não haja para aqui . & eu não estou apenas . & então quem foi ?\\\midrule
				aonde vamos nós , rosa ?  EOU  em meio caminho , minha avó .  EOU  jesus senhor , valei - me , pois que as minhas pobres pernas já estão cançadas , e parece - me que não chego ao fim da jornada . & então , vamos lá ! & o que é que eu não ? & encoste - se ao meu hombro , avósinha , que eu não estou cançada .\\\midrule
				é muita honra para mim , minha querida senhora ; estou portanto ás vossas ordens . & e então não sabe ? & se quiser a senhora . & visto isso não vos recusareis a dizer - me se estaes satisfeita com a vossa neta ?\\\midrule\midrule
				sr . hathaway .  EOU  onde está ferrante ?  EOU  como poderia saber ? & não . & pare de ser passá - lo para um pouco . & um passarinho me contou que ele deveria estar aqui .\\\midrule
				a capela parece encantadora .  EOU  não quer entrar ?  EOU  posso ? & muito bem . & não . & faça favor .\\\midrule
				você acha que eles estão apaixonados ?  EOU  não saberia dizer , gladys .  EOU  estou louca pra ver nickie ferrante . & oh ! não me entendo ! & diz - me . & não sei qual foi a pergunta de 64 milhões de dólares , mas ele tinha a resposta .\\\midrule
				não pense que foi tudo . . .  EOU  esperem aí .  EOU  fazem - me um favor ? & não , mas é preciso que não vou . & quem é um , não sei . & o quê ?\\\midrule
				ele está fazendo uma liquidação .  EOU  temos que pensar sobre isto .  EOU  onde podemos te encontrar ? & eu sei ? . . . & o que que foi com isso ? & estarei na sala de coquetel ou , claro , na sala de jantar .\\
				
				\bottomrule

			\end{tabular}
		\end{center}
		\caption{Random test samples from the Portuguese Gutenberg test set (first 5 rows), and Opensubtitles test set (last 5 rows). Responses from the GPT2 model trained on Gutenberg and Opensubtitles are compared. EOU means ``End Of Utterance''.}
	\end{table*}
	
	\begin{table*}[h!]
		\begin{center}
			\small
			\renewcommand{\arraystretch}{1.1}
			\begin{tabular}{p{6.5cm}p{2.5cm}p{2.5cm}p{2.5cm}}
			\toprule
				\bf Input & \bf Gutenberg & \bf Opensubtitles & \bf Target \\ 
				
				\midrule
				
				sieh , sieh , rico , die sonne , wie schön ! jetzt wird ’ s sommer ; sieh , wie es glitzert auf dem see . es kann gar keinen schöneren see geben , als der ist & was ist denn das ? & unverletzlich & ja , ja , stineli , du solltest nur einmal den see sehen , den ich meine !\\\midrule
natürlich hab ich dich liebgehabt . & soll ich auch gleich an mein leben herkommen ? & du warst doch gar nicht wirklich ? & gehabt !\\\midrule
was , mögen sie in den barbierladen gehen und dort warten , bis sie an die reihe kommen ? sie können ihn ja zu sich kommen lassen . ” & nein , nein , sie kommen nicht zu mir , ” erwiderte perthes . & jetzt sofort . & wir demokraten sind nicht so vornehm , ” erwidert der rechtsanwalt .\\\midrule
es gibt ärgere dinge als gefangenschaft  EOU  du galoppierst mit verhängtem zügel .  EOU  das tempo macht leben & so laß es mir selbst sagen . & und wir müssen uns anfangen & ich will mir das heut abend in mein tagebuch schreiben .\\\midrule
hallo , das boot !  EOU  hallo , das ufer !  EOU  setzt eure yolle aus und schickt die indianerin an 's land , & jetzt nicht , was die indianerin an 's boot schickt , & guten tag . & was giebt 's ?\\\midrule\midrule
du kannst dir innerlich was zerrissen haben .  EOU  du wirst dir eines tages noch den hals brechen !  EOU  gott , birdy ! & du kannst ihn ja nicht überlassen ! & was denn ? & tu so etwas nie wieder !\\\midrule
sei still .  EOU  sollen wir ihr wasser nehmen , muad'dib ?  EOU  wir gaben kampflos auf . & ja , wir gaben kampflos , & wir gaben die ganze zeit , muad 'ib . & wir ergaben uns eurer gnade .\\\midrule
wir können da nicht hinein .  EOU  wir haben keine wahl .  EOU  nichts überlebt in einem solchen sturm . & was ist denn das für ein sturm ? & so ist es nicht . & wenn sie nicht sterben wollen , werden sie uns nicht folgen , oder ?\\\midrule
gehst du mit ihm ?  EOU  tina , darf ich etwas sagen ?  EOU  tina ? & ja , tina , ich habe mit dir gesprochen . & ich hab 's ja gekämpft . & was macht ihr denn mit dem geld ?\\\midrule
und jetzt soll ich es über nacht da stehen lassen ?  EOU  holen sie es morgen wieder ab .  EOU  das geht schon . & das geht schon , & ich bin ein richtiges ekel . & wenn irgendwas mit diesem auto passiert , dann werde ich böse , tina .\\

\bottomrule

			\end{tabular}
		\end{center}
		\caption{Random test samples from the German Gutenberg test set (first 5 rows), and Opensubtitles test set (last 5 rows). Responses from the GPT2 model trained on Gutenberg and Opensubtitles are compared. EOU means ``End Of Utterance''.}
	\end{table*}

    \end{document}